\documentclass[11pt]{article}

\usepackage[final]{acl}
\usepackage{times}
\usepackage{latexsym}
\usepackage[T1]{fontenc}
\usepackage[utf8]{inputenc}
\usepackage{microtype}
\usepackage{inconsolata}
\usepackage{graphicx}
\graphicspath{{figures/}{paper/figures/}}
\usepackage{booktabs}
\usepackage{multirow}
\usepackage{amsmath}
\usepackage{verbatim}

\newcommand{\vlabel}[1]{\textsc{#1}}

\title{More Context, Larger Models, or Moral Knowledge? A Systematic Study of Schwartz Value Detection in Political Texts}

\author{
  \textbf{Víctor Yeste\textsuperscript{1,2}}
  \and
  \textbf{Paolo Rosso\textsuperscript{1,3}}
\\
\\
  \textsuperscript{1}PRHLT Research Center, Universitat Politècnica de València, Spain \\
  \textsuperscript{2}School of Science, Engineering and Design, Universidad Europea de Valencia, Spain \\
  \textsuperscript{3}Valencian Graduate School and Research Network of Artificial Intelligence (ValgrAI) \\
  \small{
    \textbf{Correspondence:} \href{mailto:vicyesmo@upv.es}{vicyesmo@upv.es}
  }
}

\begin{document}
\maketitle

\begin{abstract}
Detecting Schwartz values in political texts is hard: cues are often implicit, and neighboring values differ by fine distinctions. Two remedies are widely assumed to help: more surrounding document text, and explicit moral knowledge. Knowledge-based retrieval has improved benchmarks elsewhere, but whether either transfers here is untested, because published systems vary context, knowledge, and model family at once. We separate these factors under matched conditions on the ValuesML/Touch{\'e} ValueEval format. The input ranges from the target sentence to a local window to the full document. Retrieval is either absent or drawn from a curated moral knowledge base, injected by early, late, or cross-attention fusion. Supervised DeBERTa-v3 encoders are compared against zero-shot LLMs from 12B to 123B. More context is not uniformly better: full-document input improves the encoders by 2.5--3.8 macro-F1 points but does not consistently help the LLMs. Retrieved knowledge helps more reliably, improving every model family and context under early fusion. A control substituting random knowledge-base entries shows the families gain differently: LLMs from relevance, encoders from exposure to the value ontology. Neither larger encoders nor larger LLMs guarantee gains, and early fusion outperforms both trainable variants. Value-sensitive NLP should evaluate context, knowledge, and model family jointly.
\end{abstract}

\section{Introduction}

Political texts argue for policies, but they also appeal to values such as
security, autonomy, tradition, equality, and care, and usually do so indirectly
\citep{feldman1988structure,goren2005party,entman1993framing,chong2007framing}:
a sentence may convey societal security through a claim about migration without
naming the value. Schwartz's theory provides a structure for such distinctions
\citep{schwartz1992universals}, and the refined 19-value taxonomy is fine-grained
enough for computational analysis \citep{schwartz2012refining}. That granularity
also makes sentence-level classification hard: values are implicit, overlapping,
and rare. Each sentence is annotated on its own, so a label is a property of that
sentence, but the cues that justify it are frequently located in the text around
it \citep{falk-lapesa-2025-mining}.

Recent work frames this as multi-label human value detection in argument and
political text
\citep{kiesel-etal-2022-identifying,kiesel-etal-2023-semeval,mirzakhmedova-etal-2024-touche23,kiesel2024touche}.
These benchmarks allow systems to be compared on a shared label space, but leave
one question unresolved: what information should a model receive when deciding
whether a sentence expresses a value? Published systems answer it differently and
vary several things at once -- how much surrounding text they supply, whether
they add external knowledge, and which model family they use -- so a reported
gain can rarely be attributed to any one choice. Settling the question requires
varying them one at a time under otherwise identical conditions.

Our results confirm that the sentence alone is often insufficient: replacing it
with the full document raises macro-F1 by .025 for DeBERTa-v3-base and .038 for
DeBERTa-v3-large (Section~\ref{sec:results}). More document text is not obviously
the remedy either, since it introduces unrelated content and lengthens inputs
that model families handle differently. Explicit moral knowledge is a different
option: rather than more text from the same document, a system can retrieve
concise definitions, annotation guidance, and contrasts among values.
Retrieval-augmented methods have improved benchmarks across many tasks
\citep{lewis2020retrieval,karpukhin-etal-2020-dense}, which is why it is worth
asking whether they transfer here. They may not: the knowledge that matters is
conceptual rather than factual, and must separate boundaries as fine as
\vlabel{Benevolence: caring} from \vlabel{Universalism: concern}.

Instruction-tuned large language models complicate the picture further. Used
zero-shot they follow label definitions given in the prompt and read long
contexts, whereas supervised encoders can be tuned on the task
\citep{brown2020language,ouyang2022training}. A useful evaluation must therefore
separate effects that are usually entangled -- context, retrieved knowledge,
model family, scale, and fusion mechanism -- which matters especially for a
socially sensitive task, where a higher overall macro-F1 can hide that some
values improved while others became worse
\citep{hovy-spruit-2016-social,blodgett-etal-2020-language}.

We present a systematic study of sentence-level Schwartz value detection in
political news text, varying four factors one at a time under matched conditions:

\begin{list}{$\bullet$}{
    \setlength{\labelwidth}{1.0em}
    \setlength{\leftmargin}{1.4em}
    \setlength{\labelsep}{0.4em}
    \setlength{\itemsep}{1pt}
    \setlength{\parsep}{0pt}
    \setlength{\topsep}{3pt}
}
    \item \textbf{Input context:} the target sentence, a local window, or the
    full document.
    \item \textbf{Moral knowledge:} no retrieval, or retrieval from a curated
    knowledge base of definitions, guidance, and contrasts.
    \item \textbf{Model family and scale:} DeBERTa-v3 encoders at base and large
    scale \citep{he2021debertav3}, and zero-shot instruction-tuned LLMs spanning
    three model sizes.
    \item \textbf{Fusion mechanism:} early fusion, late fusion, or
    cross-attention on the encoder side.
\end{list}

\noindent These correspond to four research questions, each isolating a choice
current systems make implicitly:

\begin{list}{RQ\arabic{enumi}.}{
    \usecounter{enumi}
    \setlength{\labelwidth}{2.4em}
    \setlength{\leftmargin}{3.0em}
    \setlength{\labelsep}{0.6em}
    \setlength{\itemsep}{0pt}
    \setlength{\parsep}{0pt}
}
    \item How does in-document context affect value detection? Cues often lie
    outside the target sentence, but longer inputs also carry unrelated content.
    \item Does retrieved moral knowledge help beyond document context?
    Conceptual knowledge may separate neighboring values as more text cannot.
    \item How do model family, scale, and fusion strategy mediate that
    usefulness? An input usable by a trained encoder may be unusable zero-shot.
    \item Which values benefit most? An aggregate gain may be concentrated in a
    few frequent labels, which matters for a socially sensitive task.
\end{list}

Our contribution is not a new taxonomy or model but a systematic account of when
common sources of additional information help value-sensitive NLP. We evaluate
context and retrieved knowledge under matched conditions, isolate \emph{why}
retrieval helps by separating the benefit of relevant retrieval from that of
exposure to the value ontology, compare supervised and zero-shot systems without
treating scale as a sufficient explanation, and connect aggregate results to
per-value behavior and prediction changes.

\paragraph{Terminology.} The \emph{target sentence} is the sentence whose labels
are predicted; every condition predicts labels for the same target sentence and
differs only in the text supplied around it. \emph{Moral knowledge} means the
curated knowledge-base text -- value definitions, annotation guidance, and
contrasts between neighboring values -- never the values a model itself might
hold, and \emph{retrieved moral knowledge} is the small set of entries selected
for one input. The three mechanisms for combining that text with the input --
\emph{early fusion}, \emph{late fusion} and \emph{cross-attention} -- differ in
where the retrieved text enters the model, and are defined alongside
Figure~\ref{fig:rag-fusion} in Section~\ref{sec:models}. A \emph{matched
comparison} holds model, context condition, and data split fixed so that only the
factor under test changes.

\section{Related Work}
\label{sec:related-work}

\paragraph{ValueEval systems.}
We build on work treating values as organizing principles in political judgment
and framing, and on Schwartz's refined taxonomy as a computational label space
\citep{feldman1988structure,goren2005party,schwartz1992universals,schwartz2012refining},
which the ValueEval and Touch{\'e} shared tasks apply to arguments and political
texts
\citep{kiesel-etal-2022-identifying,kiesel-etal-2023-semeval,mirzakhmedova-etal-2024-touche23,kiesel2024touche}.
Shared-task systems have used transformer encoders, label definitions,
hierarchy-aware formulations, class-token attention, and DeBERTa-style
fine-tuning
\citep{devlin-etal-2019-bert,fang-etal-2023-epicurus,tsunokake-etal-2023-hitachi,aziz-etal-2023-csecu,kandru-etal-2023-tenzin,hematian-hemati-etal-2023-sutnlp,papadopoulos-etal-2023-andronicus,honda-wilharm-2023-noam,ghahroodi-etal-2023-sina,yeste2024philo}.
Closer to our setting, recent work moves the task from whole arguments to
individual sentences, asking whether a sentence carries any value and how the 19
group into higher-order categories \citep{yeste2026higher,yeste2026single}. That
work establishes the sentence-level formulation we adopt but fixes the input the
model receives; we keep the formulation and vary the input instead.

\paragraph{LLMs and value detection.}
Human value detection is related to broader moral-language analysis, including
moral-foundation classification in political and social-media text
\citep{graham2009liberals,fulgoni-etal-2016-empirical,johnson-goldwasser-2018-classification,abdulhai-etal-2024-moral}.
Value annotations also carry systematic human and model uncertainty
\citep{falk-lapesa-2025-mining}, motivating per-value analysis rather than
macro-F1 alone. Large language models make zero-shot classification practical
\citep{brown2020language,ouyang2022training}, and a growing body of work asks
which values a model itself expresses or endorses
\citep{yao-etal-2024-value,han-etal-2025-value,rodrigues-etal-2024-beyond}. Our
question differs: we do not probe the model's own dispositions but ask whether it
can recognize values someone else expressed in a text it is given, which is a
classification problem. We therefore run the LLMs zero-shot against DeBERTa
encoders supervised on that task \citep{he2021debertav3}.

\paragraph{Knowledge-enhanced models.}
A long line of work supplies encoders with knowledge they did not acquire during
pretraining: ERNIE aligns knowledge-graph entity embeddings with token
representations \citep{zhang-etal-2019-ernie}, KnowBERT inserts
knowledge-attention layers over entity linkers \citep{peters-etal-2019-knowledge},
and KEPLER trains one encoder jointly on language modeling and knowledge
embedding \citep{wang-etal-2021-kepler}. Two things separate that work from ours.
It injects \emph{factual} knowledge, mostly entities and relations, whereas what
disambiguates a value label is conceptual -- caring versus concern, personal
versus societal security. And it modifies pretraining or architecture, whereas we
supply knowledge at the input, which is what lets the model stay fixed while only
the knowledge condition varies.

\paragraph{Context and retrieval.}
Document-aware models help when meaning is distributed across sentences
\citep{yang-etal-2016-hierarchical,pappas-popescu-belis-2017-multilingual}, but
sentence-level value detection labels one marked target rather than the whole
document, and wider context can recover implicit cues or introduce unrelated
content; we therefore compare sentence, window, and document inputs explicitly.
Retrieval-augmented models combine parametric representations with external
evidence \citep{guu2020realm,lewis2020retrieval,karpukhin-etal-2020-dense} using
dense sentence embeddings \citep{reimers-gurevych-2019-sentence}, and fusion
methods integrate that evidence at different stages
\citep{izacard-grave-2021-leveraging,dong2025decoupling}. Retrieval normally
supplies facts for answering or generating text; ours differs on both counts, as
what we retrieve is conceptual -- definitions and contrasts between neighboring
labels -- and it feeds a classifier rather than a decoder. Because the retrieval
step is identical across settings, any difference between early fusion, late
fusion, and cross-attention is attributable to the fusion mechanism alone.

\paragraph{What we adopt, and what is new.}
Our components are deliberately standard, and we state what is borrowed. The
encoder recipe follows ValueEval shared-task systems
\citep{he2021debertav3,fang-etal-2023-epicurus,tsunokake-etal-2023-hitachi};
retrieval follows dense sentence-embedding search
\citep{reimers-gurevych-2019-sentence} in the framing of
\citet{lewis2020retrieval}; and the fusion mechanisms follow
\citet{izacard-grave-2021-leveraging}. What is new is the design: a matched
protocol crossing input context with retrieved knowledge across two supervised
scales and three zero-shot families, a curated moral knowledge base, and a
control separating the benefit of relevant retrieval from that of exposure to the
ontology. The gap it fills is that existing value-detection systems fix these
choices implicitly and report one configuration, so it has not been possible to
tell which of them a reported gain depends on.

\section{Dataset and Task}
\label{sec:dataset-task}

We use the ValuesML/Touch{\'e}24-ValueEval data
\citep{kiesel-etal-2022-identifying,kiesel-etal-2023-semeval,mirzakhmedova-etal-2024-touche23,kiesel2024touche}.
Because the genre matters for how the findings should be read, we state it
explicitly. The documents are English online news articles and political
manifestos reporting on political, economic, and social affairs, typically a
headline followed by a body of 28 sentences on average (median 26, range 2--85).
They are not political speeches, parliamentary debates, or argumentative essays,
and we use ``political text'' throughout in this broader sense. Each sentence
carries a document identifier and a sentence position, which let us reconstruct
local windows and full-document context for the same target. The prediction unit
is a single target sentence; splits are document-disjoint and all systems are
evaluated on the same test sentences.

The label space is the refined Schwartz taxonomy
\citep{schwartz1992universals,schwartz2012refining}, whose 19 values and
task-facing descriptions are listed in Table~\ref{tab:schwartz-values} of
Appendix~\ref{sec:schwartz-taxonomy}. We use
it rather than Moral Foundations Theory because it is the benchmark's own label
space and is finer-grained: distinctions this paper studies, such as personal
versus societal security, have no counterpart among the moral foundations. The
released labels distinguish attained from constrained values; because our
questions concern value presence we collapse both into one binary label per
value, making the task multi-label classification in which a sentence may express
no value, one, or several.

\begin{table}[t]
\centering
\small
\setlength{\tabcolsep}{4pt}
\begin{tabular}{lrrrrr}
\toprule
Split & Docs & Sent. & Lbl./sent. & No label & $>1$ label \\
\midrule
Train & 1,603 & 44,758 & 0.58 & 48.5\% & 5.9\% \\
Val. & 523 & 14,904 & 0.58 & 49.0\% & 5.9\% \\
Test & 522 & 14,569 & 0.58 & 49.2\% & 6.2\% \\
\bottomrule
\end{tabular}
\caption{Dataset statistics after collapsing attained/constrained annotations
into value-presence labels.}
\label{tab:dataset-stats}
\end{table}

Table~\ref{tab:dataset-stats} shows the task is sparse and skewed: about half of
all sentences carry no positive label and only 6\% are multi-label, with
\vlabel{Security: societal}, \vlabel{Achievement}, and \vlabel{Conformity: rules}
most frequent and \vlabel{Humility}, \vlabel{Hedonism}, and
\vlabel{Universalism: tolerance} rarest in the test split. This is why macro-F1
is the primary metric and why per-value analysis is needed to tell whether
context and retrieved knowledge help only frequent values or also rare and
conceptually subtle ones.

\section{Knowledge Base and Retrieval}
\label{sec:kb-retrieval}

We build a compact moral knowledge base (KB) to test whether explicit value
knowledge helps sentence-level classification beyond in-document context. We
describe it in terms of the three components a knowledge-based system requires:
how the knowledge is \emph{acquired}, how it is \emph{maintained}, and how it is
\emph{transferred} to the model at prediction time.

\textbf{Acquisition.} The KB
contains 58 manually curated chunks: 19 value-definition chunks, 25 operational
guideline chunks, and 14 theory-level chunks describing contrasts or relations
among values. The definition and theory chunks are grounded in the refined
Schwartz taxonomy \citep{schwartz1992universals,schwartz2012refining}; the
guideline chunks encode task-facing distinctions that are useful for annotation,
such as separating \vlabel{Security: personal} from \vlabel{Security: societal}
or \vlabel{Benevolence: caring} from \vlabel{Universalism: concern}. The KB
holds no training or test instances, which is a deliberate choice between two
kinds of store. An \emph{instance-based} store holds labeled examples for a model
to imitate, as retrieval commonly supports in-context learning
\citep{liu-etal-2022-makes}, and is tied to the training distribution. A
\emph{conceptual} store holds the distinctions an annotator would consult, stays
small enough to inspect by hand, and is unchanged when the training set changes;
its cost is that it supplies no surface patterns, which
Section~\ref{sec:randkb-control} shows to be a real cost for the encoder.

\begin{figure*}[t]
\centering
\includegraphics[width=\textwidth]{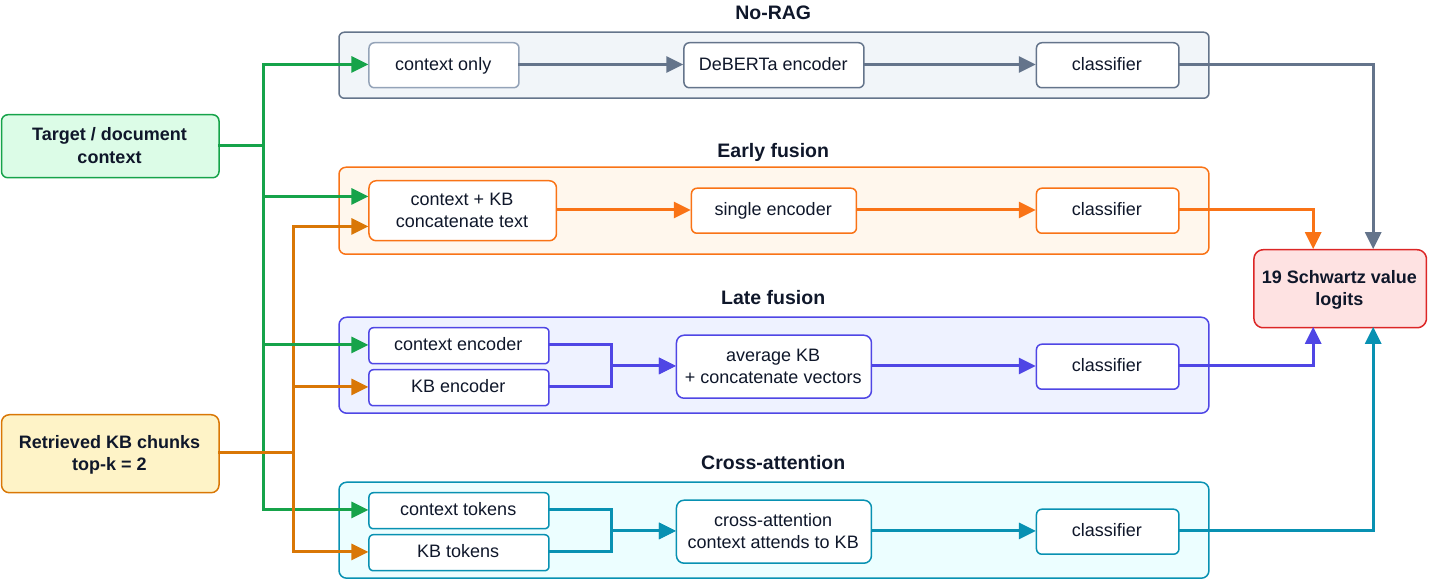}
\caption{Encoder-side RAG fusion ablation. All RAG variants use the same
retrieved KB chunks; only the fusion mechanism changes.}
\label{fig:rag-fusion}
\end{figure*}

\textbf{Maintenance.} Each chunk carries a source type (definition, guideline,
or theory) and optional value metadata, recorded for later analysis but never
used to filter retrieval, which keeps retrieval label-agnostic: the model
receives retrieved text, never gold label information. The KB is versioned as a
whole, so a result is tied to a specific KB state and chunks can be revised
without touching the retrieval or fusion code
(Appendix~\ref{sec:reproducibility}).

\textbf{Transfer.} Deciding \emph{which} knowledge is relevant to an input is
the step connecting the KB to the task. Relevance is embedding similarity: chunk
texts and the query are embedded with the same sentence-embedding model and the
nearest chunks selected by distance over a flat index
\citep{reimers-gurevych-2019-sentence,johnson2019billion}, with a fixed
top-\(k=2\) throughout. No label information, keyword rule, or value metadata
takes part (Appendix~\ref{sec:reproducibility}). What differs across conditions
is the \emph{query}: for encoder RAG it is the constructed input for the current
context condition, and for zero-shot LLM RAG it is the target sentence, with the
retrieved snippets inserted into the prompt alongside the chosen context. With
document context, retrieved text is capped by a fixed KB budget so document text
and knowledge share one input length. Retrieval is held fixed within each
comparison -- early fusion, late fusion, and cross-attention share the same KB,
embedding model, index, query construction, and \(k\) -- so differences among
them reflect fusion rather than retrieval.

Retrieved chunks match the target sentence's gold labels far more often than
chance (41.3\% against 19.0--19.3\%);
Appendix~\ref{sec:retrieval-quality} breaks this down by query type and
validates it against a judged sample.

\section{Models and Input Conditions}
\label{sec:models}

\subsection{Context Conditions}

All conditions predict labels for the same target sentence and differ only in the
text around it: the \emph{sentence} condition supplies the target alone, the
\emph{window} condition adds up to two sentences on either side (truncated at
document boundaries), and the \emph{document} condition supplies the full
document. For encoders the document condition is not built by truncating the
document, which could discard the target itself. The input is grown
\emph{outward} from the target, which is wrapped in explicit markers
(\texttt{<TGT>}\,\ldots\,\texttt{</TGT>}), adding whole sentences on either side
until the 824-token document budget is filled. The target is therefore always
present, never split by truncation, and always identifiable, wherever it sits in
the document. The same budget applies with and without retrieval, so the two
conditions see equal document text (Appendix~\ref{sec:reproducibility}). For
LLMs the target sentence is a separate prompt field, so it is never confusable
with its context.

\subsection{Supervised DeBERTa Encoders}

We use DeBERTa-v3-base and DeBERTa-v3-large \citep{he2021debertav3}, trained as
19-way multi-label classifiers with a sigmoid output per value. Training
optimizes binary cross-entropy and selects checkpoints on validation macro-F1;
predictions threshold the 19 probabilities at a validation-selected value held
fixed for test (Appendix~\ref{sec:reproducibility}). Because fine-tuning large
encoders is sensitive to initialization and data order, DeBERTa results are run
across seeds and reported as aggregates in Section~\ref{sec:results}.

\subsection{Encoder RAG Architectures}

We compare four encoder-side knowledge conditions, placed side by side in
Figure~\ref{fig:rag-fusion}, which differ only in \emph{where} the retrieved text
enters the model. In \emph{no-RAG} it never enters, and the model sees only the
selected sentence, window, or document context. In \emph{early fusion} the
retrieved chunks are concatenated with the input before encoding, so one encoder
sees context and knowledge as a single sequence. In \emph{late fusion} the
context and the chunks are encoded separately, the chunk representations are
averaged, and the two vectors are concatenated for the classifier. In
\emph{cross-attention} they are likewise encoded separately, but a
cross-attention block lets context tokens attend to knowledge tokens before
classification. The encoder, the retrieved chunks, and the classifier head are
identical in all four, which is what makes any difference attributable to the
fusion point alone rather than to retrieval or capacity.

\begin{figure}[t]
\centering
\includegraphics[width=\linewidth]{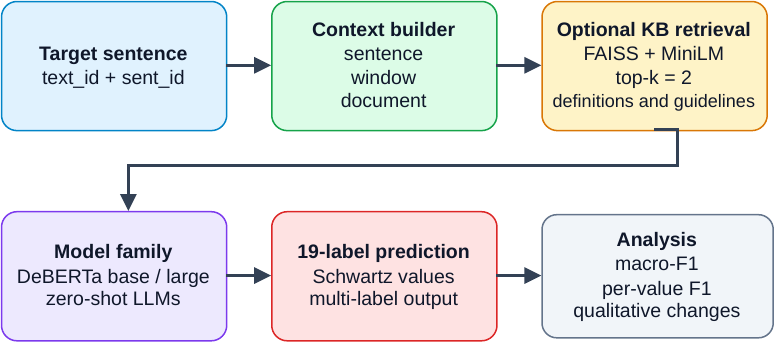}
\caption{Experiment pipeline: a fixed target sentence passes through context
construction, optional retrieval, model prediction, and analysis.}
\label{fig:study-overview}
\end{figure}

\subsection{Zero-shot LLMs}

We also evaluate instruction-tuned decoder LLMs without task-specific
fine-tuning: Gemma 3 12B IT \citep{gemmaTeam2025gemma3},
Qwen2.5-72B-Instruct \citep{yang2024qwen25technical}, and
Mistral-Large-Instruct-2407 \citep{mistralai2024mistralLargeInstruct2407}. They
serve as one representative model at each of three sizes: 12B, 72B, and 123B
parameters. We deliberately do not fine-tune them: the
question is whether instruction-tuned LLMs can detect values using only label
definitions, optional retrieved knowledge, and longer contexts supplied in the
prompt.

The prompt contains a task description, the 19 Schwartz value names with
one-line definitions, output instructions, optional retrieved KB snippets, and
the target sentence with the selected context condition. Models are instructed to
return either a comma-separated list of canonical value names or \texttt{NONE};
the full template is shown in Figure~\ref{fig:prompt-template} in
Appendix~\ref{sec:prompt-template}.
Decoding is deterministic. We parse JSON-like lists, JSON objects with a
\texttt{labels} field, and comma-, semicolon- or newline-separated text, matching
parsed strings case-insensitively against the canonical label set; unknown and
duplicate labels are dropped and \texttt{NONE} becomes the empty set.

\section{Experimental Setup}
\label{sec:experimental-setup}

Figure~\ref{fig:study-overview} traces one prediction through the study: a target
sentence is fixed, context construction places sentence, window, or document text
around it, retrieval optionally adds KB chunks, a model produces the 19 label
decisions, and analysis aggregates them per condition and per value. Each experiment is one
path through it, so a score change is attributable to the stage that varied. The main experiment crosses model
family, input context, and retrieved knowledge. DeBERTa-v3-base and
DeBERTa-v3-large are evaluated under the three context conditions of
Section~\ref{sec:models}, each with and without early-fusion RAG, giving twelve
encoder conditions; Gemma-3-12B-it, Qwen2.5-72B-Instruct and
Mistral-Large-Instruct-2407 are evaluated zero-shot under the same conditions.
For the document setting we additionally run an encoder fusion ablation over
no-RAG, early fusion, late fusion, and cross-attention at both DeBERTa scales.

\paragraph{Design choices.}
Because the study is about attribution rather than peak accuracy, every choice
was fixed on external grounds and then held constant -- the encoder family, one
open-weight instruction-tuned LLM per size band, and \(k=2\) set a priori so
retrieved text cannot crowd out document context.
Appendix~\ref{sec:design-choices} states each choice, the alternatives
considered, and a sensitivity check showing \(k\) was not tuned.

All DeBERTa models are trained on the training split, selected on validation, and
evaluated on the held-out test split, using three seeds (\(7,42,1701\)) with mean
and standard deviation reported across them to expose experimental variance
\citep{dodge-etal-2019-show}. One prediction threshold of \(0.18\) is used for every
run rather than tuned per condition. A sweep from \(0.00\) to \(1.00\) in steps
of \(0.01\) on the \emph{validation} split selects \(0.20\), only .0007 macro-F1
above \(0.18\), so the choice is not consequential in this range. Retrieval-augmented conditions use the
same FAISS index with \(k=2\), and the KB budget is capped at \(200\) tokens for
budgeted document inputs. LLM inference is deterministic, with at most \(64\)
generated tokens, and large models use 8-bit quantization when GPU memory
requires it; we return to that constraint in the limitations. Models range from
184M and 435M parameters for the two DeBERTa scales to 12B, 72B, and 123B for
Gemma, Qwen, and Mistral. Experiments ran on NVIDIA H100 80GB nodes with a budget
on the order of \(10^3\) GPU-hours. Appendix~\ref{sec:reproducibility} gives
hyperparameters and the remaining settings, and
Appendix~\ref{sec:data-code-availability} describes the released artifacts.

Macro-F1 is the primary metric because the label distribution is highly imbalanced
and the question concerns all values rather than only frequent ones; we report
micro-F1 as a secondary aggregate and per-label F1 for the value-level analysis.
For key paired contrasts we compute confidence intervals with paired bootstrap
resampling over test sentences and paired permutation tests with 2,000 iterations
\citep{dror-etal-2018-hitchhikers}. We also compare matched conditions prediction
by prediction: a \emph{prediction change} is a label decision that differs
between two conditions for the same sentence, and since each sentence carries 19
binary decisions we report change rates both per decision and per sentence,
classifying each changed sentence by whether its label set became exactly
correct, stopped being correct, or remained wrong. All tables and significance
summaries are generated from saved prediction files by the released analysis
scripts.

\section{Results}
\label{sec:results}

\subsection{RQ1: Effects of Document Context}

To isolate the effect of in-document context, Table~\ref{tab:main-results}
compares the no-RAG sentence, window, and document conditions. The clearest
pattern is that context helps supervised encoders but not zero-shot LLMs in the
same way. DeBERTa-v3-base improves from sentence-only to window and document
inputs, with document context giving the best mean macro-F1
(.285 vs. .260, a gain of +.025). DeBERTa-v3-large also benefits from
full-document input (.280 vs. .242, +.038), but the local window hurts
substantially (.207), showing that more context is not monotonically useful even
within the same encoder family. The zero-shot LLMs move in the opposite
direction: relative to sentence-only prompts, full-document prompts cost Gemma
$-$.017 and Qwen $-$.044 macro-F1, and leave Mistral slightly lower.

\begin{table*}[t]
\centering
\small
\setlength{\tabcolsep}{5pt}
\begin{tabular}{@{}lcccccc@{}}
\toprule
& \multicolumn{2}{c}{Sentence} & \multicolumn{2}{c}{Window} & \multicolumn{2}{c}{Document} \\
\cmidrule(lr){2-3}\cmidrule(lr){4-5}\cmidrule(lr){6-7}
Model & no-RAG & early RAG & no-RAG & early RAG & no-RAG & early RAG \\
\midrule
DeBERTa-B & .260$\pm$.004 & .273$\pm$.003$^{\dagger}$ & .280$\pm$.010 & .301$\pm$.005$^{\dagger}$ & .285$\pm$.013 & \textbf{.314$\pm$.008}$^{\dagger}$ \\
DeBERTa-L & .242$\pm$.004 & .258$\pm$.006$^{\dagger}$ & .207$\pm$.004 & .231$\pm$.015$^{\dagger}$ & .280$\pm$.002 & .294$\pm$.020 \\
Gemma-12B & .198 & .219$^{\dagger}$ & .194 & .217$^{\dagger}$ & .181 & .202$^{\dagger}$ \\
Qwen-72B & .215 & .241$^{\dagger}$ & .193 & .218$^{\dagger}$ & .171 & .194$^{\dagger}$ \\
Mistral-123B & .208 & .236$^{\dagger}$ & .216 & .241$^{\dagger}$ & .202 & .220$^{\dagger}$ \\
\bottomrule
\end{tabular}
\caption{Test macro-F1 for all main conditions. DeBERTa entries are
mean$\pm$standard deviation over three seeds; LLM entries are one zero-shot run.
Differences quoted in the text use full-precision scores and may differ by .001
from these rounded entries. $\dagger$ marks a RAG gain over the matched no-RAG
condition whose paired-bootstrap 95\% interval excludes zero (all three seeds for
DeBERTa, the single run for LLMs); the one unmarked RAG cell, DeBERTa-L with
document context, is significant in one seed of three. Micro-F1 and the fusion ablation are in
Appendix~\ref{sec:complete-results}.}
\label{tab:main-results}
\end{table*}

Paired bootstrap tests support the encoder-side effect: document context beats
sentence-only input for both DeBERTa scales in every seed. The window condition
is less stable, positive for base in two seeds and near-neutral in the third but
consistently negative for large. In-document
context is therefore useful when the model can learn how to use it, but adds
unrelated content and prompt length for zero-shot LLMs (RQ1).

\subsection{RQ2: Effects of Retrieved Moral Knowledge}
\label{sec:rq2}

Comparing early-fusion RAG against the matched no-RAG condition in
Table~\ref{tab:main-results}, retrieved moral knowledge improves macro-F1 for
every model family and context, modestly but consistently, from .013 to .030.
DeBERTa-v3-base averages +.021 across the three contexts; the
zero-shot LLMs average +.022 to +.025. DeBERTa-v3-large improves less, and its
document-RAG gain is more seed-sensitive than base's.

The contrast with RQ1 matters: more document text is not reliably useful to
zero-shot LLMs, but retrieved knowledge is. Gemma, Qwen, and Mistral improve
under RAG in every context, even where the longer context degraded no-RAG
performance, and paired bootstrap intervals are above zero for all LLM and all
DeBERTa-v3-base RAG contrasts. For DeBERTa-v3-large, sentence and window RAG are
consistently positive while document RAG rests on one strong seed. Retrieved
knowledge is thus a relatively reliable addition whose benefit still depends on
model scale and context format rather than acting as a uniform boost (RQ2).
Section~\ref{sec:randkb-control} isolates \emph{why} this knowledge helps, and
shows that the mechanism is not the same for the two model families.

\subsection{Random-KB Control: What Makes Retrieved Knowledge Useful}
\label{sec:randkb-control}

The RQ2 gains are consistent but modest, and two explanations fit them equally
well: the model may benefit from knowledge \emph{relevant} to the target
sentence, or merely from the presence of moral vocabulary, in which case any KB
text would do. To separate these we repeat the RAG conditions with the top
\(k=2\) \emph{retrieved} chunks replaced by two drawn uniformly at random from
the same 58-chunk KB, sampled deterministically per sentence. Chunk count, token
budget, and fusion path are held fixed, so the only variable is whether the
injected knowledge was selected for the sentence. This splits the KB gain into
\emph{priming}, random minus no-RAG, the benefit of any moral text; and
\emph{targeting}, retrieved minus random, the additional benefit of relevance.
For the supervised encoder we retrain end-to-end with random KB on all splits
(three seeds) so that it receives the same learning opportunity as retrieval; an
inference-time swap would confound the contrast with a train--test distribution
shift. The frozen zero-shot LLM needs only a prompt swap. We run the control on
DeBERTa-v3-base and on Gemma-3-12B as a second family, but not on
DeBERTa-v3-large, whose document-RAG gain is already seed-sensitive
(Section~\ref{sec:rq2}).

\begin{table}[t]
\centering
\small
\setlength{\tabcolsep}{2.5pt}
\begin{tabular}{@{}lccccc@{}}
\toprule
System & no-RAG & rand. & retr. & priming & targeting \\
\midrule
DeBERTa-B doc & .285 & .311 & .314 & +.027* & +.003 \\
\midrule
Gemma sentence & .198 & .206 & .219 & +.008* & +.013* \\
Gemma window & .194 & .194 & .217 & $-$.000 & +.024* \\
Gemma doc & .181 & .183 & .202 & +.001 & +.020* \\
\bottomrule
\end{tabular}
\caption{Random-KB control: macro-F1 without retrieval, with random KB chunks,
and with retrieved chunks, decomposed into priming and targeting. DeBERTa rows
are seed-averaged; Gemma rows are one zero-shot run. * marks a paired-bootstrap
95\% interval excluding zero -- in every seed for DeBERTa, in the single run for
Gemma.}
\label{tab:randkb-control}
\end{table}

Table~\ref{tab:randkb-control} shows the two families gain for different
reasons. For the zero-shot LLM the effect is
targeting: retrieved knowledge beats random in all three contexts (+.013, 95\% CI
[+.004, +.022]; +.024, [+.015, +.032]; +.020, [+.013, +.027]), whereas priming is
indistinguishable from zero for window (\(-\).000, \(p=.97\)) and document
(+.001, \(p=.66\)) prompts and only marginal for sentence prompts (+.008,
\(p=.037\)). Random moral text does not reliably help; only knowledge selected
for the sentence does. This also answers an objection to the zero-shot RAG
setting -- that the value definitions are already in the prompt
(Appendix~\ref{sec:prompt-template}). Precisely because they are, adding further
undifferentiated text changes nothing, and the KB's measurable contribution is
the targeted retrieval of chunks bearing on the specific sentence.

For the supervised encoder the decomposition points the other way. The total gain
is significant in every seed (+.030, +.036, +.024; all \(p<.001\)) but is
carried by priming (+.025, +.039, +.015, significant in every seed), while
targeting is significant in none (mean +.003; \(p=.31\), \(.43\), \(.06\)). A
trained encoder benefits from exposure to the definitions and guidelines in its
input, largely independently of which chunks are retrieved. The same 58 chunks
recur across the corpus, so an encoder trained with them can exploit the ontology
as a fixed part of the input rather than as sentence-specific evidence.

Retrieved moral knowledge therefore helps both families through different
mechanisms: relevance for zero-shot LLMs, which cannot adapt to the KB and must
be handed the right chunk at inference time, and ontology exposure for the
encoder, which internalizes the label semantics during training. Practically,
retrieval quality is worth investing in for prompting-based systems, whereas for
a fine-tuned encoder most of the available benefit comes from making the ontology
visible in the input at all. Reporting only the aggregate gain would have hidden
that the two families use the KB differently.

\subsection{RQ3: Model Family, Scale, and Fusion Strategy}
\label{sec:rq3-note}

Table~\ref{tab:main-results} supports the family and scale comparison; the fusion
ablation is in Appendix~\ref{sec:complete-results}. This is not a controlled
pretraining-scale study: the DeBERTa models are supervised on the task while
Gemma, Qwen, and Mistral are used zero-shot. DeBERTa-v3-base with document
early-RAG is strongest overall (.314 macro-F1), above the best zero-shot LLM
(.241). The comparison is therefore between a model given task supervision and
models given none, and that is the only claim it supports: here, a few hundred
million supervised parameters outperform 123B unsupervised ones. It does not show
that encoders are better models than Mistral-Large, nor that supervision matters
more than scale in general, since the two are not varied independently. Minimal
supervision -- few-shot examples from the training split -- could make
performance scale with size, in which case supervision and scale would reinforce
rather than compete; we did not run that condition and return to it in the
limitations. Each size is also represented by a single model from a different
family, so family and scale are confounded on the LLM side. Within what we did
vary, scale is not monotonic: DeBERTa-v3-large does not reliably improve on base,
and larger LLMs beat Gemma mainly in shorter-context RAG. Holding retrieval
fixed, early fusion is best at both scales, so the late-fusion and
cross-attention variants add complexity without improving performance (RQ3).

\subsection{RQ4: Which Values Benefit Most?}
\label{sec:rq4}

The aggregate gains in RQ1--RQ3 are not distributed uniformly across the
Schwartz taxonomy. Table~\ref{tab:rq4-values} in
Appendix~\ref{sec:per-value-results} summarizes the strongest per-value
patterns, alongside the full 19-label breakdown. Document context mainly helps values
whose interpretation depends on the surrounding social or political situation:
\vlabel{Hedonism}, \vlabel{Face}, and \vlabel{Tradition}. These labels are
difficult to infer from an isolated sentence when the sentence names an event or
stance but leaves the relevant motivation implicit.

Retrieved knowledge produces a related but distinct profile, with its largest
encoder gains on labels that require a boundary decision rather than more topic
information -- \vlabel{Benevolence: caring}, \vlabel{Stimulation}, and
\vlabel{Face}. \vlabel{Face} benefits from both interventions, consistent with
needing the social situation and the value definition at once. The long tail
stays hard: the best score anywhere for \vlabel{Humility},
\vlabel{Self-direction: thought}, and \vlabel{Conformity: interpersonal} is
below .18 F1, and support alone does not explain it. Model family shifts the
error profile rather than removing the difficulty, with the LLMs gaining most on
broad policy-facing labels and the encoders on socially situated ones (RQ4).
Appendix~\ref{sec:per-value-results} gives the full 19-label breakdown and this
analysis in detail.

\subsection{Qualitative Error Patterns}
\label{sec:qualitative-patterns}

Aggregate scores say how much context and retrieval help, not where. Comparing
matched conditions prediction by prediction shows both are targeted: replacing
the sentence with the document changes 4.3\% of label decisions for
DeBERTa-v3-base and adding retrieval changes 3.7\% (LLMs 5.1--12.2\%), while
whole label sets change far more often, 52.8\% and 46.5\% of sentences, because
one altered decision changes the set.

Direction matters more than volume. Among sentences whose label set changes when
the document replaces the sentence, 25\% become exactly correct and 18\% stop
being correct, or 1.4 corrections per new error; retrieval is noisier, at 23\%
corrected against 20\% broken, a ratio of 1.15. This is what lies behind the
modest gains of Section~\ref{sec:rq2}: retrieval does not quietly add accuracy
but revises many predictions and wins narrowly, so its benefit depends on the
retrieved chunk suiting the sentence more often than not.

Two sentences from one test document make both directions concrete (seed 42;
paraphrased per the dataset terms). In the first, a minister argues that
sustaining living standards requires immigration; sentence-only DeBERTa reads
this as \vlabel{Benevolence: caring}, but with the full document and two chunks
covering \vlabel{Security: societal} it predicts the gold
\vlabel{Security: societal}. In the second, from the same document, a
commissioner concedes that irregular arrivals must be handled better: without
retrieval the model predicts the gold \vlabel{Conformity: rules}, and with
retrieval it adds \vlabel{Security: societal} -- the very value carried by the
chunk that corrected the first sentence.

Both share one mechanism: retrieved knowledge acts as a prior toward the values
it contains, and since retrieval runs once per document input that prior reaches
every sentence, including those whose frame differs. It is right more often than
wrong, which is why scores improve; when wrong, topical relevance is mistaken for
value expression (Appendix~\ref{sec:qual-examples}).

\section{Discussion and Conclusion}
\label{sec:discussion}
\label{sec:conclusion}

The central implication is conditionality: the same added information helps or
hurts depending on the model. For supervised encoders, document context and
early-fusion retrieval are complementary -- the document recovers the political
frame while the value descriptions sharpen the label space -- but the control in
Section~\ref{sec:randkb-control} shows the second effect comes from exposure to
the ontology rather than sentence-specific retrieval, so for a trained encoder
the KB acts less like retrieval than like a structured description of the label
space. For zero-shot LLMs retrieved knowledge is more reliable than longer
prompts, and there the benefit does depend on retrieving the entries relevant to
each sentence. The results also caution against treating scale or architectural
complexity as a substitute for task design: DeBERTa-v3-large does not
consistently improve over base, larger instruction-tuned LLMs do not outperform
the supervised encoder \emph{when used zero-shot}, and late fusion and
cross-attention do not improve over early fusion. That qualification matters:
these results bound what the LLMs do without task supervision, not what they
could do with it.

Practically, they favor a conservative default: start with a supervised encoder,
choose document context deliberately, and add early-fusion knowledge when label
boundaries are ambiguous. That is cheaper than 70B--123B zero-shot LLMs, more
reproducible across seeds, and easier to inspect, since the retrieved chunks are
visible. The per-value results also show why one overall score is not enough:
systems should be judged by which values are helped, harmed, and left
unresolved.

The numbers belong to one benchmark, but the design transfers: a randomized
control like that in Section~\ref{sec:randkb-control} separates gains due to
\emph{relevance} from gains due to mere \emph{presence}, and the answer says
where to invest. The question outlives RAG: a longer window still leaves open
whether a model uses the right part of it.

\section*{Limitations}

This study covers one benchmark and one broad genre, news reporting on political
and social affairs together with political manifestos, so the conclusions may not
transfer to social media, parliamentary debates, speeches, or longer
argumentative essays. It also uses the English task formulation and English KB
entries, leaving multilingual transfer open.

Comparisons across model families carry an asymmetry: the encoders are trained
and evaluated on splits of one corpus, so part of their advantage is adaptation
to this distribution, whereas the zero-shot LLMs receive none, and
in-distribution and out-of-distribution behaviour can diverge sharply in NLP
\citep{yuan2023revisiting}. The clean test would train the encoders elsewhere,
but that needs a label-space mapping that is not one-to-one:
\citet{kiesel-etal-2022-identifying} uses argument-level categories and
\citet{yao-etal-2024-value} targets the values a model expresses, not a text's.
Cross-corpus generalisation is future work.

The KB is fixed and manually built, so results depend on its coverage and
wording, and different chunking or retrieval models could change RAG behaviour.
Retrieval uses a fixed top-$k$, and document context is supplied verbatim within
a token budget rather than condensed; prompt and context compression
\citep{li-etal-2025-prompt} would test whether the encoder's document gain
reflects information or merely length.

The LLM experiments are zero-shot, which contrasts task-specific encoders with
general-purpose models but does not establish an upper bound: chain-of-thought,
the selection-based prompting of \citet{zhu2025eavit}, few-shot prompting,
calibration, or fine-tuning could all change the ranking. Prompting was held
fixed to avoid confounding the factors the study isolates, so our LLM scores are
a floor. The most informative missing condition is few-shot prompting with training-split
examples: if performance then scaled with size, supervision and scale would
reinforce each other rather than substitute. Large-model runs may
also need quantization or multi-GPU execution, which can affect outputs.

Finally, the architecture ablations are not exhaustive. Late fusion and
cross-attention may need further tuning, and our cross-attention variant fuses
only at the final encoder layer, so the simplest mechanism winning here does not
show that richer fusion cannot help. Per-value results for rare labels, notably
\vlabel{Humility}, rest on small support and are best read as broad patterns.

\section*{Ethical Considerations}

Value detection is sensitive because outputs can be read as claims about
political actors or communities. Our intended use is aggregate research analysis
of textual framing, not automated judgment of individual beliefs or political
legitimacy; predictions are made under uncertainty and are analytical signals
requiring human interpretation.

Misclassification can harm if labels are used to profile speakers, rank
viewpoints, or support moderation and surveillance, especially for minority or
contested positions where framing is subtle. We therefore discourage
individual-level profiling, automated moderation, and high-stakes decisions;
transparent, auditable research is the appropriate setting. The KB's wording can
also shape behaviour, so we document its construction and make retrievals
inspectable.

Better value detection also has constructive uses: studying how public debate is
framed, letting value-aware assistants recognise what a user cares about rather
than inferring it from demographics, and giving alignment research a handle on
whether outputs reflect a text's values. All require the same precondition:
inspectable predictions, judged per value.

\section*{Acknowledgments}

The authors used GPT-5.5 and Claude Opus 5 for language polishing, structural editing,
code organization and result-extraction scripts. The relevance ratings in the
retrieval audit of Appendix~\ref{sec:retrieval-quality} were likewise
model-drafted against the rubric stated there. The authors reviewed and edited
all generated text, code, and outputs, and are responsible for all claims,
analyses, and citations.

This work received support from ELENI: Estrategia Local En Narrativas
Inteligente, a project funded by CDTI (IDI-20260067), and Multimodal and
Multisensor data-centric AI models (ANNOTATE-MULTI2), grant PID2024-156022OB-C32
funded by MICIU/AEI/10.13039/501100011033 and by ERDF/EU.

\bibliography{custom,anthology-custom}

\appendix

\section{Data and Code Availability}
\label{sec:data-code-availability}

The benchmark texts are distributed by the shared-task organizers under their
own access conditions, and we do not redistribute the raw corpus texts. We
release the source code, configuration files for all training and inference
runs, prompt templates, retrieval KB files, Slurm scripts, environment
documentation, artifact documentation, and analysis scripts used to build the tables and
qualitative examples.\footnote{https://github.com/VictorMYeste/human-value-detection-context-rag}

We also release aggregate result files, tuned thresholds where applicable, and
prediction files in a form permitted by the dataset license. Where a prediction or
qualitative-analysis artifact would contain restricted text, we instead provide
the script and configuration needed to regenerate it after obtaining the
official dataset. Fine-tuned DeBERTa checkpoints and Hugging Face model bundles
are released where permitted by the base-model and dataset terms.\footnote{https://huggingface.co/VictorYeste/value-context-rag-deberta-v3-base-doc-rag} For large instruction-tuned LLMs, we release only configurations,
prompts, and derived outputs rather than redistributing model weights. Given
access to the official data under its original terms, these artifacts reproduce
all results reported in this paper.

\section{Schwartz 19-Value Taxonomy}
\label{sec:schwartz-taxonomy}

\begin{figure*}[t]
\centering
\includegraphics[width=0.86\textwidth]{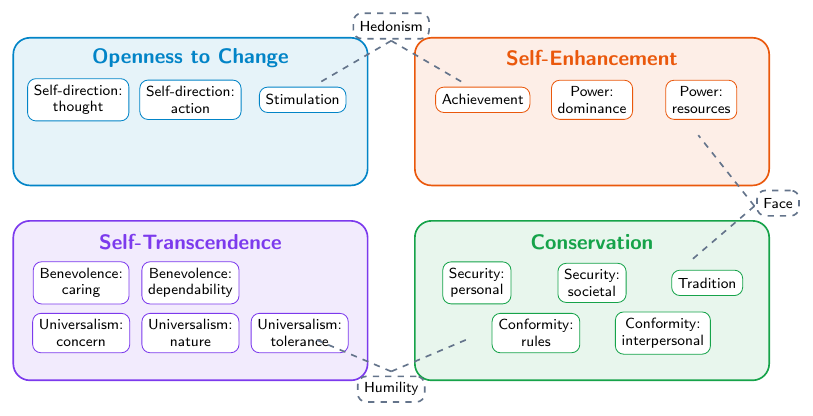}
\caption{Reader orientation only: this map is not used by any model, and no
experiment consumes the circular structure. Compact map of the refined
Schwartz 19-value taxonomy used
as the label space. Dashed labels indicate boundary values in the motivational
continuum.}
\label{fig:schwartz-taxonomy}
\end{figure*}

Figure~\ref{fig:schwartz-taxonomy} gives a compact orientation map of the 19
refined Schwartz values used in the task. The higher-order regions are shown for
interpretability only; all experiments predict the 19 values independently as
binary multi-label targets. Table~\ref{tab:schwartz-values} lists the same 19
values with the task-facing descriptions used during annotation.

\begin{table*}[t]
\centering
\small
\setlength{\tabcolsep}{5pt}
\renewcommand{\arraystretch}{1.08}
\begin{tabular}{p{0.28\textwidth}p{0.64\textwidth}}
\toprule
Value & Short description \\
\midrule
Self-direction: thought & Freedom to cultivate one's own ideas and abilities. \\
Self-direction: action & Freedom to determine one's own actions. \\
Stimulation & Excitement, novelty, and change. \\
Hedonism & Pleasure and sensuous gratification. \\
Achievement & Success according to social standards. \\
Power: dominance & Power through exercising control over people. \\
Power: resources & Power through control of material and social resources. \\
Face & Maintaining one's public image and avoiding humiliation. \\
Security: personal & Safety in one's immediate environment. \\
Security: societal & Safety and stability in the wider society. \\
Tradition & Maintaining and preserving cultural, family, or religious traditions. \\
Conformity: rules & Compliance with rules, laws, and formal obligations. \\
Conformity: interpersonal & Avoidance of upsetting or harming other people. \\
Humility & Recognizing one's insignificance in the larger scheme of things. \\
Benevolence: caring & Devotion to the welfare of in-group members. \\
Benevolence: dependability & Being a reliable and trustworthy member of the in-group. \\
Universalism: concern & Commitment to equality, justice, and protection for all people. \\
Universalism: nature & Preservation of the natural environment. \\
Universalism: tolerance & Acceptance and understanding of those who are different from oneself. \\
\bottomrule
\end{tabular}
\caption{Task-facing descriptions of the 19 Schwartz value labels.}
\label{tab:schwartz-values}
\end{table*}

\section{Zero-shot LLM Prompt Template}
\label{sec:prompt-template}

All zero-shot LLM conditions use the same prompt structure. Retrieval-augmented
conditions insert the optional \texttt{EXTERNAL KNOWLEDGE} block before the
sentence, window, or document body. Model-specific chat templates, when present,
wrap this user prompt without changing its text. Figure~\ref{fig:prompt-template}
shows the exact template.

\begin{figure*}[t]
\centering
\footnotesize
\begin{minipage}{0.98\textwidth}
\begin{verbatim}
TASK:
You are a classifier for human values in sentences. Given a TARGET
SENTENCE and its context, identify which Schwartz values are present.

SCHWARTZ VALUE DEFINITIONS:
- Self-direction: thought: Freedom to cultivate one's own ideas and abilities
- Self-direction: action: Freedom to determine one's own actions
- Stimulation: Excitement, novelty, and change
- Hedonism: Pleasure and sensuous gratification
- Achievement: Success according to social standards
- Power: dominance: Power through exercising control over people
- Power: resources: Power through control of material and social resources
- Face: Maintaining one's public image and avoiding humiliation
- Security: personal: Safety in one's immediate environment
- Security: societal: Safety and stability in the wider society
- Tradition: Maintaining and preserving cultural, family, or religious traditions
- Conformity: rules: Compliance with rules, laws, and formal obligations
- Conformity: interpersonal: Avoidance of upsetting or harming other people
- Humility: Recognising one's insignificance in the larger scheme of things
- Benevolence: caring: Devotion to the welfare of in-group members
- Benevolence: dependability: Being a reliable and trustworthy member of the in-group
- Universalism: concern: Commitment to equality, justice, and protection for all people
- Universalism: nature: Preservation of the natural environment
- Universalism: tolerance: Acceptance and understanding of those who are different from oneself

INSTRUCTIONS:
- Output a comma-separated list of value names from the definitions above.
- If no values are present, output: NONE
- Output only the list (or NONE), no extra text.

[Optional for RAG]
EXTERNAL KNOWLEDGE:
- <retrieved KB chunk 1>
- <retrieved KB chunk 2>

[Sentence condition]
TARGET SENTENCE:
<target sentence>

[Window condition]
CONTEXT WINDOW:
<local context window>

TARGET SENTENCE:
<target sentence>

[Document condition]
DOCUMENT:
<document context>

TARGET SENTENCE:
<target sentence>
\end{verbatim}
\end{minipage}
\caption{Zero-shot LLM prompt template. The optional external-knowledge block is
included only for RAG conditions; exactly one of the sentence, window, or
document bodies is used for each input condition.}
\label{fig:prompt-template}
\end{figure*}

\section{Design Choices and Alternatives}
\label{sec:design-choices}

Because the point of the study is attribution rather than peak accuracy, we fixed
each choice on external grounds and then held it constant. DeBERTa-v3 is the
encoder family used by the strongest ValueEval shared-task systems
\citep{fang-etal-2023-epicurus,tsunokake-etal-2023-hitachi}, and its base and
large variants vary scale within one family; a weaker encoder would lower every
number without changing which factor is responsible for a gain. The LLMs are
open-weight and instruction-tuned so that the runs are reproducible, one per size
band. The alternative -- one family at three sizes -- would isolate scale more
cleanly, but no open family spanned roughly 12B to 123B at comparable
instruction quality when the study was run, so we chose scale coverage over
family purity and treat the resulting confound as a limit on the scale claim
(Section~\ref{sec:rq3-note}). The three context conditions are the natural
gradations of in-document context, with the window fixed at two sentences on
either side. Retrieval uses \(k=2\) chunks, chosen a priori so that retrieved
text cannot crowd out document context within the 1{,}024-token limit rather
than by tuning: a single-seed check on the sentence condition varying \(k\) from
1 to 7 gives macro-F1 between .269 and .285 with no monotone trend, and two
independent trainings of the same configuration differ by .006, so most of that
spread is training noise. We therefore did not tune \(k\), and the reported RAG
gains are not the product of a retrieval search.

\section{Knowledge Base Contents}
\label{sec:kb-examples}

Table~\ref{tab:kb-examples} shows one chunk of each type. The three are chosen
around a single distinction -- personal versus societal security -- to show how
the types complement one another: the definition states what the value means, the
guideline gives the annotator operational cues and explicit exclusions, and the
theory chunk contrasts the value with its nearest neighbor.

\begin{table*}[t]
\centering
\small
\setlength{\tabcolsep}{5pt}
\renewcommand{\arraystretch}{1.1}
\begin{tabular}{@{}p{0.10\textwidth}p{0.14\textwidth}p{0.70\textwidth}@{}}
\toprule
Type & Values & Text (abridged) \\
\midrule
definition & \vlabel{Security: societal} &
Security: societal concerns the safety, stability, and functioning of the wider
society. It is about social order, national security, strong institutions, and
protection from external or internal threats to the community or country. \\
guideline & \vlabel{Security: societal} &
Use this label when the sentence concerns safety, order or stability for society,
the country or clearly defined social groups as a whole, including economic
stability and national defence. Do not use it when the focus is primarily on
caring for vulnerable groups across borders (\vlabel{Universalism: concern}) or on
individual experiences of danger (\vlabel{Security: personal}). \\
theory & \vlabel{Security: personal}, \vlabel{Security: societal} &
When a sentence focuses on keeping families, neighbourhoods or individuals safe,
Security: personal is more likely; when it emphasises defending the country,
protecting borders or maintaining law and order in society, Security: societal is
more likely. \\
\bottomrule
\end{tabular}
\caption{One knowledge-base chunk of each type. The KB holds 58 chunks in total:
19 value definitions, 25 operational guidelines, and 14 theory-level contrasts.}
\label{tab:kb-examples}
\end{table*}

A worked retrieval makes the mechanism concrete. Test sentence
\texttt{EN\_M\_032:1} sets out the remit of a national defence force:
protecting the state against armed attack, together with maritime and fisheries
security (paraphrased per the dataset terms; gold label
\vlabel{Security: societal}). The two nearest chunks by embedding
distance are the \vlabel{Security: societal} definition above and the
\vlabel{Security: personal} definition. The retrieval is therefore on-topic but
not precise: it returns the correct value together with its nearest competitor,
which is the pattern the audit in Appendix~\ref{sec:retrieval-quality} quantifies
-- most retrievals are useful, a minority are decisive, and the errors are
confusions between neighboring values rather than unrelated material.

\section{Effect of the Generation Cap}
\label{sec:generation-cap}

Zero-shot outputs are capped at 64 generated tokens, which raises the question of
whether label lists are being cut off. We audited every saved generation: 87,414
per model family, over the six context-by-retrieval conditions. Outputs that
parsed to neither a canonical value name nor \texttt{NONE} account for 0.18\% of
Gemma generations, 0.001\% of Qwen, and 0.014\% of Mistral. Outputs carrying a
truncation signature -- a final comma-separated item that is a strict prefix of a
canonical label -- account for 0.50\%, 0.07\%, and 0.001\% respectively. The
longest output produced by any model is 54 words, well inside the cap.

The cap therefore affects at most half a percent of predictions in the worst
family, and the affected cases are concentrated in the smallest model. Label
recall is not meaningfully constrained by it: gold label sets are small, with
87.8\% of labeled sentences carrying a single value and the largest carrying
seven, so the cap would have to truncate lists that models rarely produce.

\section{Retrieval Quality}
\label{sec:retrieval-quality}

We validate retrieval in two ways. The automatic proxy asks whether the value
metadata attached to the two retrieved chunks intersects the gold labels of the
target sentence, over the 7,402 test sentences that carry at least one label.
Because the retrieval query is the constructed input, the proxy varies with the
context condition: 41.3\% for a target-sentence query, 36.4\% for a window query,
and 32.7\% for a full-document query, against 19.0\% when chunks are drawn at
random from the same KB for the encoder and 19.3\% for the zero-shot LLM. Every
zero-shot LLM condition uses the target sentence as the query and therefore
shares the 41.3\% rate regardless of how much context is placed in the prompt.

Two things follow. Retrieval is well above chance in every condition, by a factor
of roughly 1.7 to 2.1. But its precision degrades monotonically as the query
broadens, which means the encoder's best configuration -- document context with
early-fusion RAG -- is also the one whose retrieved chunks are least specific to
the sentence being labeled. That is consistent with the random-KB control in
Section~\ref{sec:randkb-control}, where the encoder's gain proved to come from
exposure to the ontology rather than from per-sentence targeting.

The proxy is a lower bound on relevance, because a chunk can be pertinent to a
sentence without its metadata naming exactly the gold label, and it says nothing
about whether a retrieved definition is the \emph{right} one to disambiguate a
borderline case. We therefore validated it against a judged sample.

We drew 100 labeled test sentences at random from the sentence-query condition,
with their two retrieved chunks, giving 200 judgements. Each chunk was rated 2 if
it bears directly on assigning a gold label for that sentence, 1 if it covers an
adjacent value or a general annotation guideline that applies, and 0 if it has no
bearing. Ratings were drafted by a large language model against this rubric and
then reviewed and verified by the authors; the sample, the ratings, and the
script that produces both are released with the artifact.

Of the 200 chunks, 23.5\% were rated decisive, 42.0\% related and 34.5\%
unrelated, so 65.5\% carried some useful information. At the sentence level,
39.0\% of sentences received at least one decisive chunk, against the 41.3\%
estimated by the automatic proxy for the same condition -- the two agree to
within 2.3 points, which is the main result of this check.

The disagreements are one-sided and explain the gap. Of the 49 chunks the proxy
counts as hits, 46 were rated decisive and none unrelated, so the proxy almost
never credits a chunk that a reader would reject. Of the 151 it counts as misses,
however, 54\% were still rated useful. Part of that is structural: 36 of the 200
retrieved chunks are annotation guidelines carrying no value metadata -- when to
prefer NONE, how to treat reported speech, how to read a policy proposal -- which
the proxy cannot credit by construction, and 24 of those were rated useful. The
proxy is therefore precise but conservative, and the true rate of useful
retrieval is higher than the numbers in the main text suggest.

\section{Reproducibility Details}
\label{sec:reproducibility}

Table~\ref{tab:reproducibility-details} summarizes the main settings needed to
reproduce the reported experiments, assuming access to the official benchmark
data under its original terms.

An encoder document input has the form
\texttt{$\ldots$ prior sentence.\ <TGT>target sentence</TGT>\ next
sentence.\ $\ldots$}, grown outward from the target until the document budget is
filled. That budget is 824 of the 1{,}024 available tokens in both the no-RAG and
the RAG condition; the remaining 200 tokens are reserved for retrieved knowledge
and are simply unused without retrieval. This keeps the two document conditions
matched on document text, at the cost of not filling the encoder window in the
no-RAG case.

\begin{table*}[t]
\centering
\small
\setlength{\tabcolsep}{5pt}
\renewcommand{\arraystretch}{1.05}
\begin{tabular}{@{}p{0.24\textwidth}p{0.68\textwidth}@{}}
\toprule
Component & Setting \\
\midrule
Prediction unit & Target sentence identified by \texttt{text\_id} and
\texttt{sent\_id}; train, validation, and test splits are document-disjoint. \\
Labels & Nineteen refined Schwartz values; attained and constrained annotations
are collapsed into one binary value-presence label. \\
Supervised seeds & DeBERTa runs use seeds \(7\), \(42\), and \(1701\); tables
report mean and standard deviation across seeds. \\
Thresholding & The sigmoid decision threshold is selected on validation and
fixed at \(0.18\) for test evaluation. \\
DeBERTa-v3-base & Learning rate \(1{\times}10^{-5}\), weight decay \(0.15\),
batch size \(8\), gradient accumulation \(2\), max length \(1024\). \\
DeBERTa-v3-large & Learning rate \(3{\times}10^{-6}\), weight decay \(0.1\),
batch size \(16\), gradient accumulation \(2\), max length \(1024\), gradient
checkpointing, fp32 training. \\
Training control & Up to \(20\) epochs with early stopping, maximum gradient norm
\(1.0\), checkpoint selection by validation macro-F1. \\
Software & Implemented with PyTorch \citep{paszke2019pytorch} and HuggingFace
Transformers \citep{wolf-etal-2020-transformers}; metrics use scikit-learn
\citep{pedregosa2011scikit}. Package versions and launch scripts are provided
with the released artifact. \\
KB records & Each chunk is a JSONL record with a unique identifier, a source
type (\texttt{definition}, \texttt{guidelines}, or \texttt{theory}), the chunk
text, and optional value metadata, recorded for analysis only. \\
Classifier head & HuggingFace sequence-classification interface with
\texttt{problem\_type=multi\_label\_classification}; binary cross-entropy with
logits over 19 sigmoid outputs. \\
Retrieval & \texttt{sentence-transformers/all-MiniLM-L6-v2} embeddings,
FAISS \texttt{IndexFlatL2}, normalized chunk vectors, fixed top-\(k=2\). \\
Compute & NVIDIA H100 80GB GPU nodes; encoder and Gemma jobs used one GPU,
Qwen-72B jobs used two GPUs, and Mistral-123B jobs used four GPUs. The reported
runs used approximately \(10^3\) allocated GPU-hours including resumed runs. \\
Document RAG budget & Retrieved KB text is capped at \(200\) tokens; the
remaining encoder budget is assigned to document context around the target. \\
LLM decoding & Deterministic decoding with temperature \(0\), top-\(p=1\), and
maximum \(64\) generated tokens. \\
Large LLM loading & Automatic device placement; 8-bit quantization is used when
required by GPU memory. \\
Analysis scripts & Final aggregate tables, per-value tables, prediction changes,
and qualitative bundles are generated from saved prediction files with the
released analysis scripts. \\
\bottomrule
\end{tabular}
\caption{Compact reproducibility summary for the main experiments.}
\label{tab:reproducibility-details}
\end{table*}

\section{Complete Test Results}
\label{sec:complete-results}

Table~\ref{tab:complete-test-results} reports the full set of aggregated test
results used in the main analysis. DeBERTa rows report mean$\pm$standard
deviation across three fine-tuning seeds. Zero-shot LLM rows report one
completed inference run per condition.

\begin{table*}[t]
\centering
\small
\setlength{\tabcolsep}{4pt}
\renewcommand{\arraystretch}{1.02}
\begin{tabular}{@{}llccc@{}}
\toprule
Model & Context & RAG/fusion & Macro-F1 & Micro-F1 \\
\midrule
DeBERTa-B & sent. & none & .260$\pm$.004 & .324$\pm$.010 \\
DeBERTa-B & sent. & early & .273$\pm$.003 & .338$\pm$.007 \\
DeBERTa-B & window & none & .280$\pm$.010 & .337$\pm$.007 \\
DeBERTa-B & window & early & .301$\pm$.005 & .364$\pm$.001 \\
DeBERTa-B & doc & none & .285$\pm$.013 & .346$\pm$.012 \\
DeBERTa-B & doc & early & \textbf{.314$\pm$.008} & \textbf{.369$\pm$.010} \\
DeBERTa-B & doc & late & .294$\pm$.011 & .350$\pm$.008 \\
DeBERTa-B & doc & cross & .301$\pm$.007 & .368$\pm$.013 \\
\midrule
DeBERTa-L & sent. & none & .242$\pm$.004 & .308$\pm$.008 \\
DeBERTa-L & sent. & early & .258$\pm$.006 & .332$\pm$.004 \\
DeBERTa-L & window & none & .207$\pm$.004 & .272$\pm$.007 \\
DeBERTa-L & window & early & .231$\pm$.015 & .291$\pm$.015 \\
DeBERTa-L & doc & none & .280$\pm$.002 & .340$\pm$.006 \\
DeBERTa-L & doc & early & .294$\pm$.020 & .349$\pm$.022 \\
DeBERTa-L & doc & late & .280$\pm$.004 & .347$\pm$.002 \\
DeBERTa-L & doc & cross & .280$\pm$.006 & .348$\pm$.004 \\
\midrule
Gemma-12B & sent. & none & .198 & .224 \\
Gemma-12B & sent. & early & .219 & .247 \\
Gemma-12B & window & none & .194 & .209 \\
Gemma-12B & window & early & .217 & .233 \\
Gemma-12B & doc & none & .181 & .201 \\
Gemma-12B & doc & early & .202 & .223 \\
\midrule
Qwen-72B & sent. & none & .215 & .232 \\
Qwen-72B & sent. & early & .241 & .264 \\
Qwen-72B & window & none & .193 & .199 \\
Qwen-72B & window & early & .218 & .233 \\
Qwen-72B & doc & none & .171 & .175 \\
Qwen-72B & doc & early & .194 & .209 \\
\midrule
Mistral-123B & sent. & none & .208 & .225 \\
Mistral-123B & sent. & early & .236 & .256 \\
Mistral-123B & window & none & .216 & .232 \\
Mistral-123B & window & early & .241 & .258 \\
Mistral-123B & doc & none & .202 & .211 \\
Mistral-123B & doc & early & .220 & .234 \\
\bottomrule
\end{tabular}
\caption{Complete aggregated test results. \emph{Early} denotes early-fusion
RAG. \emph{Late} and \emph{cross} are the encoder-only document RAG fusion
variants.}
\label{tab:complete-test-results}
\end{table*}

\section{Per-Value Results}
\label{sec:per-value-results}

Two tables support RQ4. Table~\ref{tab:rq4-values} summarizes the strongest
patterns compactly: the values helped most by document context, those helped most
by retrieved knowledge, and those that remain hardest for every system.
Table~\ref{tab:per-value-results} then gives the full 19-label evidence behind
it, with the document and knowledge columns computed on DeBERTa-v3-base so that
they match the compact summary, and the best-F1 column reporting the highest mean
per-value F1 observed across all model, context, and RAG conditions.

\begin{table*}[t]
\centering
\small
\setlength{\tabcolsep}{3pt}
\begin{tabular}{@{}p{0.16\textwidth}p{0.78\textwidth}@{}}
\toprule
Pattern & Values with largest effects \\
\midrule
Document context &
\vlabel{Hedonism} +.100; \vlabel{Face} +.089; \vlabel{Tradition} +.086 \\
Retrieved KB &
\vlabel{Benevolence: caring} +.064; \vlabel{Stimulation} +.062; \vlabel{Face} +.060 \\
Hard labels &
\vlabel{Conformity: interpersonal} .133; \vlabel{Self-direction: thought} .156; \vlabel{Humility} .179 \\
\bottomrule
\end{tabular}
\caption{Compact per-value patterns on the test set. The first two rows report
DeBERTa-v3-base $\Delta$F1; hard labels report the best observed F1 across all
tested systems and input conditions.}
\label{tab:rq4-values}
\end{table*}

Retrieved knowledge helps most for \vlabel{Benevolence: caring},
\vlabel{Stimulation}, \vlabel{Face}, \vlabel{Security: personal}, and
\vlabel{Universalism: tolerance}, which suggests it supports conceptual boundary
decisions rather than adding topical context: these are labels for which the same
sentence can plausibly be read through more than one value frame.
\vlabel{Humility} is the rarest test label, but frequency is not the whole
story, since \vlabel{Self-direction: thought} and
\vlabel{Conformity: interpersonal} have more support and still require subtle
distinctions between ideas, actions, and social harm. Among the zero-shot LLMs,
the largest document-level RAG gain for all three models is for
\vlabel{Power: resources}, with \vlabel{Universalism: concern} and
\vlabel{Conformity: rules} also recurring, so the larger instruction-tuned
models use retrieved value descriptions most effectively for broad policy-facing
categories.

\begin{table*}[t]
\centering
\small
\setlength{\tabcolsep}{4pt}
\renewcommand{\arraystretch}{1.02}
\begin{tabular}{@{}p{0.27\textwidth}rrrrp{0.18\textwidth}@{}}
\toprule
Value & Gold & Doc $\Delta$ & KB $\Delta$ & Best F1 & Best setting \\
\midrule
Self-direction: thought & 171 & -.003 & +.028 & .156 & Qwen sent. none \\
Self-direction: action & 512 & +.073 & +.033 & .212 & D-B doc early \\
Stimulation & 371 & +.018 & +.062 & .294 & D-B doc early \\
Hedonism & 125 & +.100 & +.034 & .341 & Mistral window none \\
Achievement & 911 & +.042 & +.038 & .395 & D-B doc early \\
Power: dominance & 631 & +.054 & +.004 & .349 & D-B doc cross \\
Power: resources & 805 & +.038 & +.024 & .423 & D-B doc early \\
Face & 267 & +.089 & +.060 & .257 & D-B doc early \\
Security: personal & 352 & +.048 & +.047 & .375 & D-B window early \\
Security: societal & 1151 & +.031 & +.026 & .450 & D-B doc cross \\
Tradition & 196 & +.086 & +.001 & .458 & D-B window early \\
Conformity: rules & 911 & +.033 & +.005 & .447 & D-B doc cross \\
Conformity: interpersonal & 195 & +.043 & +.038 & .133 & D-L doc cross \\
Humility & 30 & +.019 & -.005 & .179 & Qwen sent. early \\
Benevolence: caring & 324 & +.046 & +.064 & .292 & D-B doc early \\
Benevolence: dependability & 288 & +.080 & +.022 & .287 & D-B doc early \\
Universalism: concern & 735 & +.021 & +.022 & .432 & D-B window early \\
Universalism: nature & 293 & +.017 & +.019 & .594 & D-L doc early \\
Universalism: tolerance & 171 & +.072 & +.043 & .262 & D-B doc early \\
\bottomrule
\end{tabular}
\caption{Full per-value test results supporting RQ4. Doc $\Delta$ is
DeBERTa-v3-base document no-RAG minus sentence no-RAG. KB $\Delta$ is
DeBERTa-v3-base document early-RAG minus document no-RAG. D-B and D-L denote
DeBERTa-v3-base and DeBERTa-v3-large.}
\label{tab:per-value-results}
\end{table*}

\section{Qualitative Examples}
\label{sec:qual-examples}

Table~\ref{tab:qual-examples} reports representative examples used to support
the qualitative analysis in Section~\ref{sec:qualitative-patterns}. One further
case illustrates improved abstention: for a property report stating that house
prices rose by under 1\% in a year, sentence-only DeBERTa predicts
\vlabel{Achievement} from the language of rising prices, whereas with document
context and retrieved chunks covering unrelated values the model abstains,
matching the empty gold set (seed 42). The rows are sampled
from the qualitative bundles and prediction-change summaries generated by the
released analysis scripts. To avoid reproducing full document contexts, the
table gives paraphrased target descriptions and sentence identifiers; full
contexts can be regenerated from the official data with the released scripts.

\begin{table*}[t]
\centering
\small
\setlength{\tabcolsep}{3pt}
\renewcommand{\arraystretch}{1.08}
\begin{tabular}{@{}p{0.11\textwidth}p{0.22\textwidth}p{0.28\textwidth}p{0.16\textwidth}p{0.17\textwidth}@{}}
\toprule
Example & Pattern & Paraphrased target description & Gold & Prediction change \\
\midrule
\texttt{EN\_002:11} &
Context/RAG selects the more specific societal-security frame. &
A ministerial statement linking immigration to maintaining national living
standards. &
\vlabel{Security: societal} &
\vlabel{Benevolence: caring} $\rightarrow$ \vlabel{Security: societal} \\
\texttt{EN\_021:2} &
Retrieved guidance prevents over-annotating a descriptive market fact. &
A factual report that house prices were nearly unchanged over the year. &
None &
\vlabel{Achievement} $\rightarrow$ None \\
\texttt{EN\_002:8} &
LLM RAG separates societal risk from broad concern for others. &
A sentence describing national demographic trends as worrying. &
\vlabel{Security: societal} &
\vlabel{Universalism: concern} $\rightarrow$ \vlabel{Security: societal} \\
\texttt{EN\_008:1} &
LLM RAG abstains on a value-neutral political headline. &
A headline reporting a political claim that the economy needs a reset. &
None &
\vlabel{Achievement}, \vlabel{Security: societal}, \vlabel{Self-direction:
action} $\rightarrow$ None \\
\texttt{TR\_062:11} &
Failure: price and cost cues trigger resource/care labels but miss societal
stability. &
A sentence explaining that a cost increase affected producers and consumers. &
\vlabel{Achievement}; \vlabel{Security: societal} &
\vlabel{Power: resources} $\rightarrow$ \vlabel{Benevolence: caring};
\vlabel{Power: resources} \\
\texttt{TR\_059:18} &
Failure: the gold labels depend on document-level diplomatic motivation. &
A diplomatic statement setting a near-term bilateral trade target. &
\vlabel{Benevolence: caring}; \vlabel{Self-direction: action} &
\vlabel{Achievement}; \vlabel{Power: resources} $\rightarrow$ None \\
\bottomrule
\end{tabular}
\caption{Representative qualitative examples. To respect the dataset usage
agreement, target sentences are paraphrased rather than quoted verbatim, and
each row shows the baseline prediction followed by the prediction under the
changed condition. Rows are grouped as two supervised DeBERTa conditions, two
zero-shot LLM RAG conditions, and two failure cases.}
\label{tab:qual-examples}
\end{table*}

\end{document}